\documentclass[runningheads,a4paper]{llncs}

\usepackage{amssymb}
\usepackage{graphicx}
\usepackage{amsmath}

\usepackage{url}
\urldef{\mailsa}\path|{napoli, pappalardo, tramontana}@dmi.unict.it|

\newcommand{\keywords}[1]{\par\addvspace\baselineskip
  \noindent\keywordname\enspace\ignorespaces#1}

\begin{document}

\mainmatter  

\title{A hybrid neuro--wavelet predictor for QoS control and
  stability}

\markboth{A hybrid neuro--wavelet predictor for QoS control and
  stability -- PREPRINT}%
{Shell \MakeLowercase{\textit{et al.}}: Bare Demo of
IEEEtran.cls for Journals}

 \begin{titlepage}
 \begin{center}
 {\Large \sc PREPRINT VERSION\\}
  \vspace{5mm}
{\huge A Hybrid Neuro--Wavelet Predictor for QoS Control and Stability\\}
 \vspace{10mm}
 {\Large C. Napoli, G. Pappalardo, and E. Tramontana\\}
 \vspace{5mm}
{\Large \sc PUBLISHED ON: \bf AI*IA 2013: Advances in Artificial Intelligence\\}
 \end{center}
 \vspace{5mm}
 {\Large \sc BIBITEX: \\}
 
@incollection\{\\
year=\{2013\},\\
isbn=\{978-3-319-03523-9\},\\
booktitle=\{AI*IA 2013: Advances in Artificial Intelligence\},\\
volume=\{8249\},\\
series=\{Lecture Notes in Computer Science\},\\
editor=\{Baldoni, Matteo and Baroglio, Cristina and Boella, Guido and Micalizio, Roberto\},\\
doi=\{10.1007/978-3-319-03524-6\_45\},\\
title=\{A Hybrid Neuro--Wavelet Predictor for QoS Control and Stability\},\\
url=\{http://dx.doi.org/10.1007/978-3-319-03524-6\_45\},\\
publisher=\{Springer International Publishing\},\\
keywords=\{Neural networks; wavelet analysis; QoS; adaptive systems\},\\
author=\{Napoli, Christian and Pappalardo, Giuseppe and Tramontana, Emiliano\},\\
pages=\{527-538\}\\
\}

 \vspace{5mm}
 \begin{center}
Published version copyright \copyright~2013 SPRINGER \\
\vspace{5mm}
UPLOADED UNDER SELF-ARCHIVING POLICIES\\
NO COPYRIGHT INFRINGEMENT INTENDED \\
 \end{center}
\end{titlepage}

\title{A hybrid neuro--wavelet predictor for QoS control and
  stability}

\titlerunning{A hybrid neuro--wavelet predictor for QoS control and
  stability}

\author{Christian Napoli \and Giuseppe Pappalardo \and Emiliano
  Tramontana}

\authorrunning{Napoli C., Pappalardo G., Tramontana E.}

\institute{Dipartimento di Matematica e Informatica, University of
  Catania\\
  Viale Andrea Doria 6, 95125 Catania, Italy\\
  \mailsa}

\maketitle

\begin{abstract}
  For distributed systems to properly react to peaks of requests,
  their adaptation activities would benefit from the estimation of the
  amount of requests.  This paper proposes a solution to produce a
  short-term forecast based on data characterising user behaviour of
  online services.  We use \emph{wavelet analysis}, providing
  compression and denoising on the observed time series of the amount
  of past user requests; and a \emph{recurrent neural network} trained
  with observed data and designed so as to provide well-timed
  estimations of future requests.  The said ensemble has the ability
  to predict the amount of future user requests with a root mean
  squared error below 0.06\%.  Thanks to prediction, advance resource
  provision can be performed for the duration of a request peak and
  for just the right amount of resources, hence avoiding
  over-provisioning and associated costs. Moreover, reliable provision
  lets users enjoy a level of availability of services unaffected by
  load variations.

\end{abstract}

\section{Introduction}

General public internet usage has become an essential everyday matter,
e.g.\ widespread use of social--networks and mass communication media,
public administration, home banking, etc.  Moreover, the request for
stable and continuous internet services has reached a pressing
priority.
Generally, the solutions employed by service providers to adapt
server-side resources on-the-fly have been based on \emph{content
  adaptation} and \emph{differentiated service}
strategies~\cite{Abdelzaher02,ccpe06,BannoMPT10b,ccpe13,Kaqudai}.
However, such strategies could over-deteriorate the service during
load peaks.  Moreover, for ensuring a minimum level of quality (QoS),
even when sudden variations on the number of requests arise, a large
number of (over-provisioned) resources is often used, hence incurring
into relevant costs and wasted resources for a considerable time
interval.
Some approaches guarantee a minimum QoS level once the connection has
been established (i.e.\ non-adaptive multimedia
services)~\cite{Epstein95,Novelli07}, or by using bandwidth adaption
algorithms (i.e.\ adaptive wireless services)~\cite{Kwon99}.
However, such solutions are still liable to an effective loss of QoS
for end users, when we consider content quality or availability, and,
in the worst case, even denial of services (DoS).  As an alternative
to deterministic scheduling~\cite{Kang98} or genetic
algorithms~\cite{Sherif00}, neural networks have been used within the
area of a multi-objective optimisation problems.  In~\cite{Ahn04}
authors provide dynamic admission control while preserving QoS by
using hardware-based Hopefield neural networks~\cite{Zadeh10}.

When autoregressive moving average models (ARMA), and related
generalisations, are used, the underlying assumption, even for the
study forecasting host load~\cite{Dinda00}, makes them inappropriate
for predictions in non-linear systems exhibiting high levels of
variations.
Artificial neural networks have been used in several ways to provide
an accurate model of the QoS evolution over time.  Linear regression
models have been applied with the support of neural networks
in~\cite{Islam12}, however without some proper mechanisms, such as
time delays or feedback, it is still not possible to dynamically
follow the evolution of the extended time series.
Machine learning approaches have also been used~\cite{Powers05},
however such approaches were not designed for on-the-fly adaptation,
and are unable to give advantages with respect to user perceived
responsiveness~\cite{Schechter98}.
For the above approaches in which the amount of connection requests is
unknown, to avoid overloading the server-side, only load balancing and
admission control policies have been used.  Still when the amount of
requests overcomes the available resources, service usability
worsening or denial of service cannot be avoided.
On the other hand, when more resources can be dynamically allocated,
since the amount of the required resources is unknown in advance, it
often results in over-provisioning, with negative effects on
management and related cost.

This paper investigates the use of Second Generation Wavelet Analysis
and Recurrent Neural Network (RNN) to predict over time the amount of
connection requests for a service, by using a hybrid wavelet recurrent
neural network (WRNN).
Recurrent neural networks have been proven powerful enough to gain
advantages from the statistical properties of time series.
In our experiments, the proposed WRNN has been used to analyse data
for the Page view statistics Wikimedia(TM) project, produced by Domas
Mituzas and released under Creative Common License\footnote{%
  \textsf{See dumps.wikimedia.org/other/pagecounts-raw} }.  Wavelet
analysis has been used in order to reduce data redundancies so as to
obtain a representation that can express their intrinsic structure,
while the neural networks have been used to have the complexity of
non-linear data correlational perform data prediction.
Thereby, a relatively accurate forecast of the connection request time
series can be achieved even when load peaks arise.
The estimated result is fundamental for a management service that
performs resource preallocation on demand. The precision of our
estimates allows just the right amount of resources to be used.

The rest of this paper is structured as follows.  In
section~\ref{wavelet} the background on wavelet theory is given.
Section~\ref{secondgen} describes second generation wavelets and how
their properties are useful for the proposed forecast.
Section~\ref{wrnn} provides the proposed neural network configuration.
Section~\ref{experiments} reports on the performed experiments and
results. Finally, Section~\ref{conclusions} draws our conclusions.

\section{The basis of Wavelet Theory}
\label{wavelet}

This work builds on wavelets and neural networks to model the main
characteristics of user behaviour for an online service.
Wavelet decomposition is a powerful analysis tool for physical and
dynamic phenomena that reduces the data redundancies and yields a
compact representation expressing the intrinsic structure of a
phenomenon.  In fact, the main advantage when using wavelet
decomposition is the ability to pack the energy signature of a signal
or a time series, and then to express relevant data as a few non-zero
coefficients.  This characteristic has been proven very useful to
optimise the performances of neural networks~\cite{Gupta04}.
Like sine and cosine for Fourier transforms, a wavelet decomposition
uses functions, i.e.\ wavelets, to express a function as a particular
expansion of coefficients in the wavelet domain.  Once a mother
wavelet has been chosen, it is possible, as explained in the
following, to create new wavelets by dilates and shifts of the mother
wavelet. Such novel generated wavelets, if chosen with certain
criteria, eventually form a Riesz basis of the Hilbert space
$L^2(\mathbb{R})$ of square integrable functions.
Such criteria are at the basis of \emph{wavelet theory} and come from
the concept of multiresolution analysis of a signal, also called
multiscale approximation.  When a dynamic model can be expressed as a
time-dependent signal, i.e.\ described by a function in
$L^2(\mathbb{R})$, then it is possible to obtain a multiresolution
analysis of such a signal.  For the space $L^2(\mathbb{R})$ such an
approximation consists in an increasing sequence of closed subspaces
which approximate, with a greater amount of details, the space
$L^2(\mathbb{R})$, eventually reaching a complete representation of
$L^2(\mathbb{R})$ itself.  A complete description of multiresolution
analysis and the relation with wavelet theory can be found
in~\cite{Mallat09}.

One-dimensional decomposition wavelets of order $n$ for a signal
$s(t)$ give a new representation of the signal itself in an
$n$-dimensional multiresolution domain of coefficients plus a certain
residual coarse representation of the signal in time. For any discrete
time step $\tau$ then, the corresponding $M$ order wavelet
decomposition $\mathbf{W} s(\tau)$ of the signal $s(\tau)$ will be
given by the vector
\begin{equation}
\mathbf{W}s(\tau) = \left [ d_1(\tau), d_2(\tau) , \ldots , d_M(\tau)
,a_M(\tau) \right ] ~~~~\forall ~ \tau \in \{\tau_1,\tau_2,\ldots,\tau_N\}
\label{eq:utau1}
\end{equation}
where $d_1$ is the most detailed multiresolution approximation of the
series, and $d_M$ the least detailed, and $a_M$ is the residual
signal.  Such coefficients express some intrinsic time-energy feature
of a signal, i.e.\ features of a time series, while removing
redundancies, and offering a well suited representation, which, as
described in Section~\ref{secondgen}, we give as inputs for a neural
network.

It is now possible to give a more rigorous definition of a
wavelet. Let us take into account a multiresolution decomposition of
$L^2(\mathbb{R})$
\begin{equation*}
\varnothing \subset V_0 \subset \ldots \subset V_j \subset V_{j+1} \subset
\ldots \subset L^2(\mathbb{R})
\end{equation*}

If we call $W_j$ the orthogonal complement $V_j$, then it is possible to define a
wavelet as a function $\psi(x)$ if the set of $\{\psi(x-l)|l \in \mathbb{Z}\}$
is a Riesz basis of $W_0$ and also meets the following two constraints:
\begin{equation}
\int_{-\infty}^{+\infty} \psi(x) dx = 0
\label{eq:p4}
\end{equation}
and
\begin{equation*}
|| \psi(x) ||^2 = \int_{-\infty}^{+\infty} \psi(x)\psi^{*}(x) dx = 1
\end{equation*}
If the wavelet is also an element of $V_0$ then it exists a sequence $\{g_k\}$ such
that
\begin{equation*}
\psi(x) = 2 \sum\limits_{k \in \mathbb{Z}}
g_k \psi(2x-l)
\end{equation*}
then the set of functions $\{ \psi_{j,l} | j,l
\in \mathbb{Z}\}$ is now a Riesz basis of $L^2(\mathbb{R})$.
It follows that a wavelet function can be used to define an Hilbert basis, that
is a complete system, for the Hilbert space $L^2(\mathbb{R})$. In this case, the
Hilbert basis is constructed as the family of functions $\{ \psi_{j,l} |
j,l \in \mathbb{Z}\}$ by means of dilation and translation of a mother
wavelet function $\psi$ so that $\psi_{j,l} =
\sqrt{2^j}\psi(2^j x - l)$.  Hence,  given a function $f \in L^2(\mathbb{R})$
it is possible to obtain the following decomposition
\begin{equation}
f(x)=\sum\limits_{j,l \in \mathbb{Z}} \langle f | \psi_{j,l}
\rangle=\sum\limits_{j,l \in \mathbb{Z}} d_{j,l} \psi_{j,l}(x)
\label{eq:p7}
\end{equation}
where $d_{j,l}$ are called wavelet coefficients of the given function
$f$ in the wavelet basis given by the inner product of $\psi_{j,l}$.
Likewise, a projection on the space $V_j$ is given by
\begin{equation*}
\mathbb{P}_j f(x) = \sum\limits_{i} \langle f | \varphi_{i,j} \rangle
\varphi_{i,j}(x)
\end{equation*}
where $\varphi_{i,j}$ are called dual scaling functions. When the
basis wavelet functions coincide with their duals the basis is
orthogonal. Choosing a wavelet basis for the multiresolution analysis
corresponds to selecting the dilation and shift coefficients.  In this
way, by performing the decomposition we obtain the $\{d_i|a_M\}$
coefficients sets of~\eqref{eq:utau1}.  From now on we will refer to
the described schema as first generation wavelets, whilst
Section~\ref{secondgen} describes second generation wavelets.

For the present work, we adopted Biorthogonal wavelet decomposition
(this wavelet family is described in~\cite{Mallat09}), for which
symmetrical decomposition and exact reconstruction are possible with
finite impulse response (FIR) filters~\cite{Rabiner75}.

\section{Second generation Wavelets with RNNs}
\label{secondgen}

A multiresolution analysis like the one described in
Section~\ref{wavelet} can be realised by conjugate wavelet filter
banks that decompose the signal, and act similarly to a low/high pass
filter couple~\cite{Sweldens95}.
An advanced solution has been devised to obtain the same decomposition
using a lifting and updating procedure. This procedure, named
\emph{second generation wavelet}, takes advantage of the properties of
multiresolution wavelet analysis starting from a very simple set-up
and gradually building up a more complex multiresolution decomposition
to have some specific properties.  The lifting procedure is made of a
space-domain superposition of biorthogonal wavelets developed
in~\cite{Sweldens98}.

The construction is performed by an iterative procedure called
\emph{lifting and prediction}.  Lifting consists of splitting the
whole time series $x[\tau]$ into two disjoint subsets $x_e[\tau]$ and
$x_o[\tau]$, for the even and odd positions, respectively; whereas
prediction consists of generating a set of coefficients $d[\tau]$
representing the error of extrapolation of time series $x_o[\tau]$
from series $x_e[\tau]$.  Then, an update operation combines the
subsets $x_e[\tau]$ and $d[\tau]$ in a subset $a[\tau]$ so that

\begin{equation}
\left \{
\begin{array}{rcl}
d[\tau]&=&x_o[\tau] - \mathbf{P} x_e[\tau] \\
a[\tau]&=&x_e[\tau] + \mathbf{U} d[\tau]
\label{eq:predictor}
\end{array}
\right .
\end{equation}
where $\mathbf{P}$ is the prediction operator, and $\mathbf{U}$ is the
update operator.  Eventually, one cycle for the above procedure
creates a complete set of discrete wavelet transforms (DWT) and the
relative coefficients $d[\tau]$ and $a[\tau]$. It follows that
\begin{equation}
\left \{
\begin{array}{rcl}
x_e[\tau]&=&a[\tau] - \mathbf{U} d[\tau]\\
x_o[\tau]&=&d[\tau] + \mathbf{P} x_e[\tau]
\end{array}
\right .
\end{equation}

The said construction yields discrete wavelet filters that preserve
only a certain number $N$ of low order polynomials in the time
series. Having such low-order polynomials, in turn, makes it possible
to apply non-linear predictors without affecting the $a[\tau]$
coefficients, which provide a coarse approximation of the time series
itself.

A neural network can be build to perform such a construction, i.e.\ a
neural network would act as an inverse second generation wavelet
transform.
In~\cite{Capizzi12}, a neural network with a rich representation of
past outputs like a fully connected recurrent neural network (RNN),
known as the Williams-Zipser network or Nonlinear Autoregressive
network with eXogenous inputs (NARX)~\cite{Williams89a}, has been
proven able to generalise and reproduce the behaviour of $\mathbf{P}$
and $\mathbf{U}$ operators, and to structure itself to behave as an
optimal discrete wavelet filter.
Moreover, for such a kind of RNNs, when applied to the prediction and
modelling of stochastic phenomena, like the considered behaviour of
users, which lead to a variable number of access requests in time,
real time recurrent learning (RTRL) has been proven to be very
effective.
A complete description of RTRL algorithm, NARX and RNNs can be found
in~\cite{Williams89b,Haykin09}.

RTRL has been used to train the RNN and such a trained RNN achieves
the ability to perform lifting stages, hence the matching of the time
series dynamics at the corresponding wavelet scale.
This construction brings the possibility to match non-polynomial and
nonlinear signal structures in an optimised straightforward
N-dimension means square problem~\cite{Mandic01}.
NARX networks have been proven able to use the intrinsic features of
time series in order to predict the following values of the
series~\cite{Napoli10a}.  One of a class of transfer functions for the
RNN has to be chosen to approximate the input-output behaviour in the
most appropriate manner.
For phenomena having a deterministic dynamic behaviour, the relative
time series at a given time point can be modelled as a functional of a
certain amount of previous time steps.  In such cases, the used model
should have some internal memory to store and update context
information~\cite{Lapedes86}.  This is achieved by feeding the RNN
with a delayed version of past data, commonly referred as time delayed
inputs~\cite{Connor94}.

\section{Proposed setup for WRNN predictor}
\label{wrnn}

As stated in Section~\ref{secondgen}, it would be desired to have a
neural network able to perform the wavelet transform as in a recursive
lifting procedure. For this, we could use a mother wavelet as transfer
function, however mother wavelets lack of some elementary properties
needed by a proper transfer function, such as e.g.\ the absence of
local minima and a sufficient graded and scaled
response~\cite{Gupta04}.
This leads us to look for a close enough substitute to approximate the
properties of a mother wavelet without affecting the functionalities
of the network itself.  The function classes that more closely
approximate a mother waveform have to be found among the Radial Basis
Functions (RBFs) that are good enough as transfer functions and
partially approximate half of a mother waveform. It is indeed possible
to properly scale and shift a couple of RBFs to obtain a mother
wavelet.  If we define an RBF function as an
$f:[-1,1]\rightarrow\mathbb{R}$ then we could dilate and scale it to
obtain a new function
\begin{equation}
\tilde{f}(x+2l)=\left\{
\begin{array}{rcl}
+f(2x+1)&~~&x \in [-1,0)\\
-f(2x-1)&~~&x \in (0,+1]
\end{array}
\right . ~~~~\forall~ l \in \mathbb{Z}
\label{eq:tildef}
\end{equation}
With such a definition, starting from the properties of the RBF, it is
then possible to
verify the following
\begin{equation}
\int_{2h+1}^{2k+1} \tilde{f}(x) dx = 0 ~~~~\forall~ (h,k) \in \mathbb{Z}^2 : h <
k 
\label{eq:tildint}
\end{equation}

\begin{figure}[t]
\centering
\begin{minipage}[t]{0.22\textwidth}
\centering
\includegraphics[width=.99\textwidth]{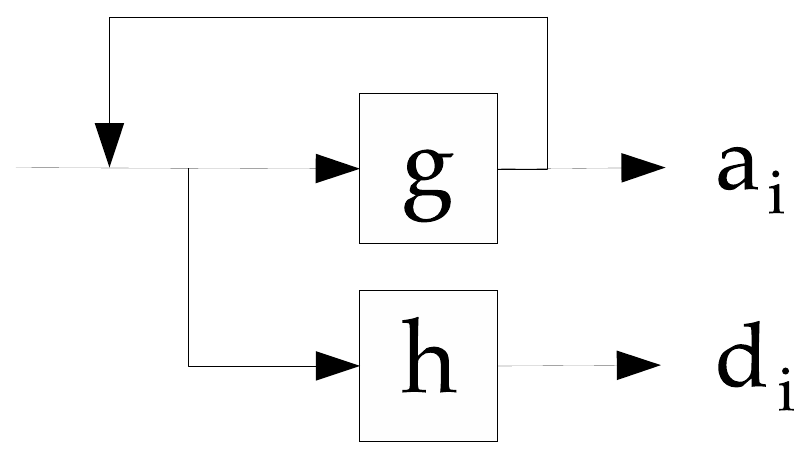}
\caption{Filter logic}
\label{fig:filtri}
\end{minipage}
\begin{minipage}[t]{0.77\textwidth}
\centering
\includegraphics[width=.99\textwidth]{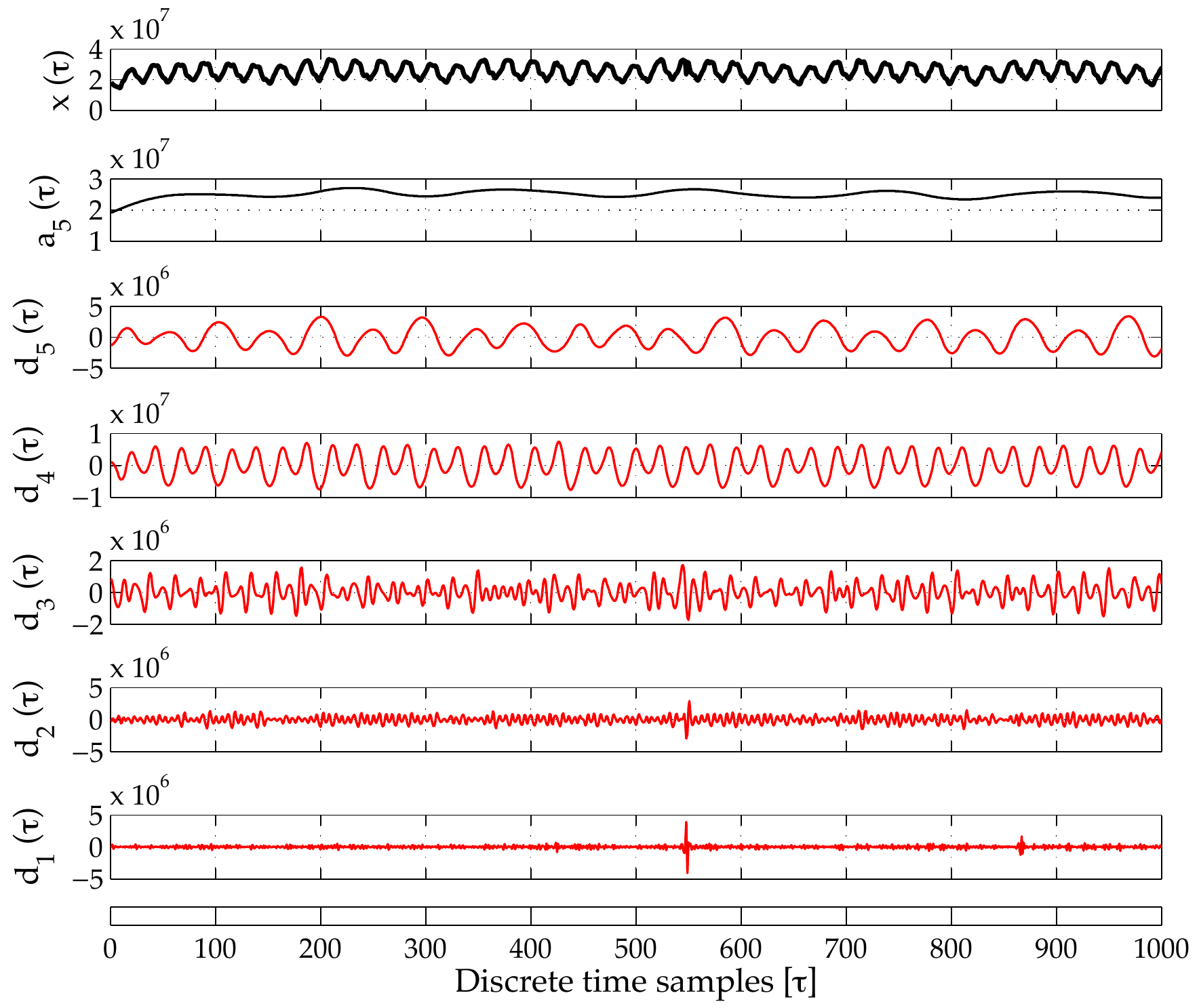}
\caption{5-level biorthogonal 3.7 wavelet decomposition}
\label{fig:decomp}
\end{minipage}
\end{figure}

Starting from \eqref{eq:tildef} and \eqref{eq:tildint} it is possible
to verify the equations \eqref{eq:p4} to \eqref{eq:p7} for the chosen
$\tilde{f}$ which we can now call a mother wavelet.  The chosen mother
wavelet is a composition of two RBF transfer functions that are
realised by the proposed neural network to obtain the properties of a
wavelet transform.  The proposed WRNN has two hidden layers with RBF
transfer function.

For this work, the initial dataset was a time series representing
access requests coming from users.  We call this series $x(\tau)$,
where $\tau$ is the discrete time step of the data, sampled with
intervals of one hour.  A biorthogonal wavelet decomposition of the
time series has been computed to obtain the correct input set for the
WRNN as required by the devised architecture.  This decomposition has
been achieved by applying the wavelet transform as a recursive couple
of conjugate filters~(Figure~\ref{fig:filtri}) in such a way that the
$i$-esime recursion $\hat{W}_i$ produces, for any time step of the
series, a set of coefficients $d_i$ and residuals $a_i$, and so that

\begin{equation}
\hat{W}_i [a_{i-1}(\tau)] = [d_i(\tau) ,  a_i(\tau) ] ~~~~~ \forall~ i \in [1,M]\cap\mathbb{N}
\label{eq:xtau}
\end{equation}
where we intend $a_0(\tau)=x(\tau)$. The input set can then be
represented as an $N \times (M+1)$ matrix of $N$ time steps of a $M$
level wavelet decomposition (see Figure~\ref{fig:decomp}), where the
$\tau$-esime row represents the $\tau$-esime time step as the
decomposition
\begin{equation}
\mathbf{u}(\tau) = \left [ d_1(\tau), d_2(\tau) , \ldots , d_M(\tau)
,a_M(\tau) \right ] 
\label{eq:utau}
\end{equation}

Each row of this dataset is given as input value to the $M$ input
neurons of the proposed WRNN. The properties of this network make it
possible, starting from an input at a time step $\tau_n$, to predict
the effective number of access requests at a time step
$\tau_{n+r}$. In this way the WRNN acts like a functional

\begin{equation}
\hat{N}[\mathbf{u}(\tau_n)] = x(\tau_{n+r})
\label{eq:Nfunctional}
\end{equation}
where $r$ is the number of time steps of forecast in the future.

\begin{figure}[t]
  \centering
  \includegraphics[width=.72\textwidth]{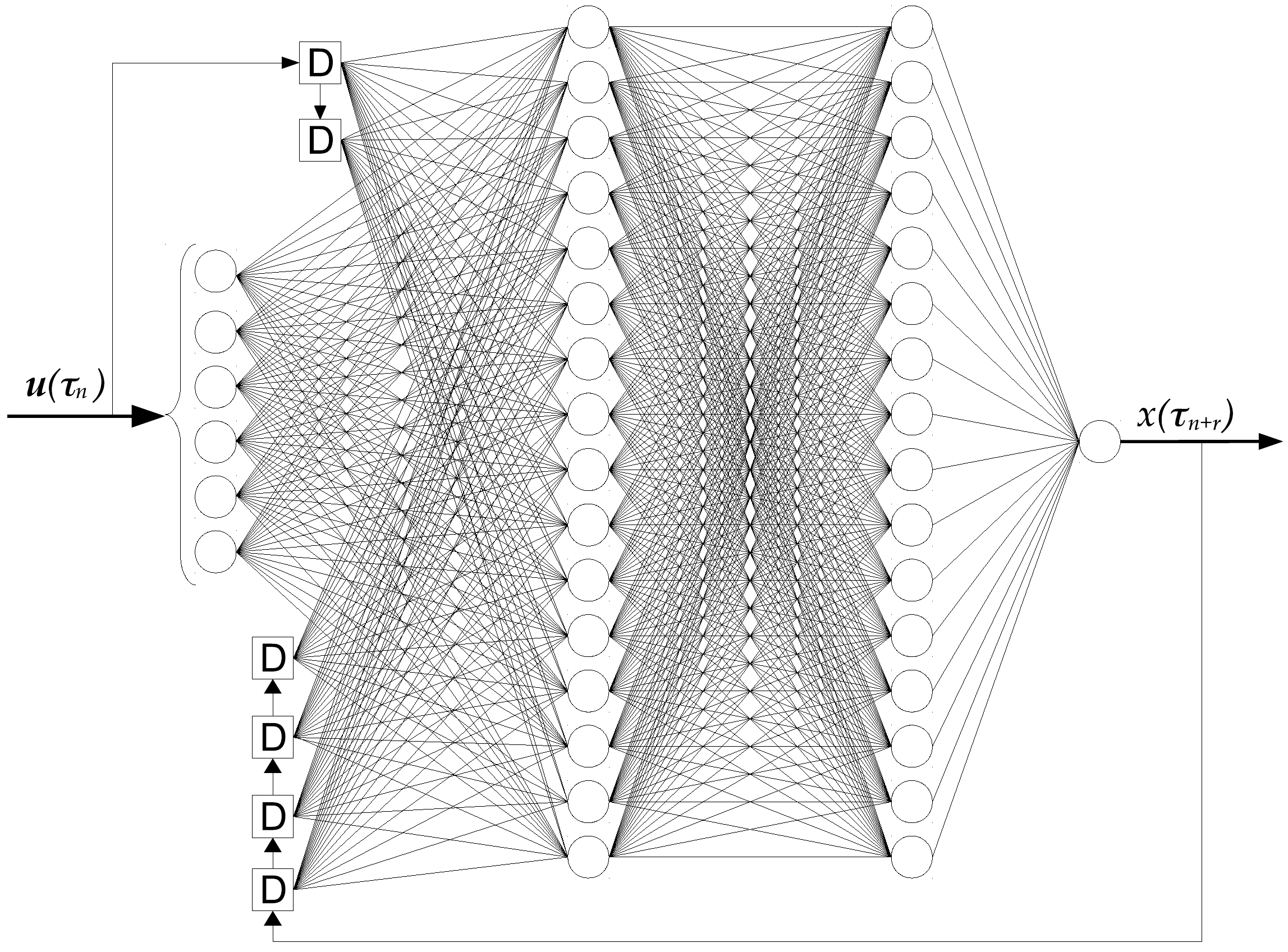}
  \caption{Devised neural netwok}
  \label{fig:nnet}
\end{figure}

\section{Experimental setup}
\label{experiments}

For this work, a 4-level wavelet decomposition has been selected that
properly characterises data under analysis.  Therefore, the devised
WRNN~(Figure~\ref{fig:nnet}) uses a 5 neuron input layer (one for each
level detail coefficient $d_i$ and one for the residual $a_5$). This
WRNN architecture presents two hidden layers with sixteen neurons each
and realises a radial basis function as explained in
Section~\ref{wrnn}.

Inputs are given to the WRNN in the following form:
\begin{itemize}
\item The wavelet decomposition of the time series
  $\mathbf{u}(\tau_n)$ for time step $\tau_n$ 
\item The previous delayed decompositions $\mathbf{u}(\tau_{n-1})$ and
  $\mathbf{u}(\tau_{n-2})$ 
\item The last four delayed outputs $x(\tau_{n+r})$ predicted by the WRNN
\end{itemize}

Delays and feedback are obtained by using the relative delay lines and
operators (D).  These feedback lines provide the WRNN with internal
memory, hence the modelling abilities for dynamic phenomena.

An accurate study has shown that the biorthogonal wavelet
decomposition optimally approximates and denoises the time series
under analysis.  Such a wavelet family is in good agreement with
previous optimal results, obtained by the authors, for the
decomposition of other physical phenomena.  In fact, such a
decomposition splits a phenomenon in a superposition of mutual and
concurrent predominant processes with a characteristic time-energy
signature.  For stochastically-driven processes, such as stellar
oscillations~\cite{ICAISC}, renewable energy and systems load
\cite{Capizzi10,Bonanno12}, and for a large category of complex
systems \cite{Napoli10b}, wavelet decomposition gives a unique and
compact representation of the leading features for a time-variant
phenomenon.

\begin{figure}[t]
\centering
\begin{minipage}[t]{0.49\textwidth}
\centering
\includegraphics[width=.99\textwidth]{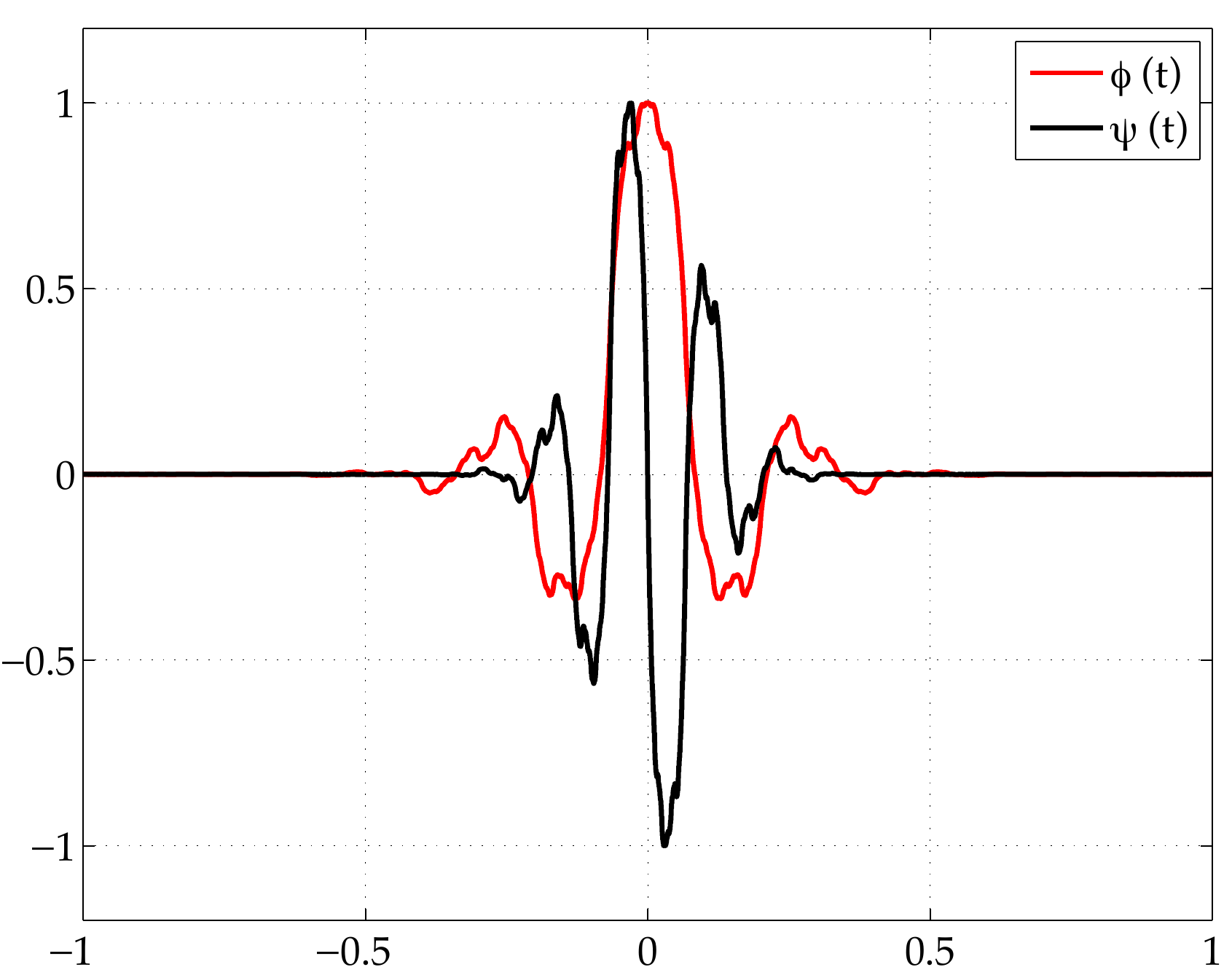}
\caption{Biorthogonal wavelet 3.7 decomposition ($\psi$) and
  scaling ($\phi$) functions} 
\label{fig:wletf}
\end{minipage}
\begin{minipage}[t]{0.49\textwidth}
\centering
\includegraphics[width=.99\textwidth]{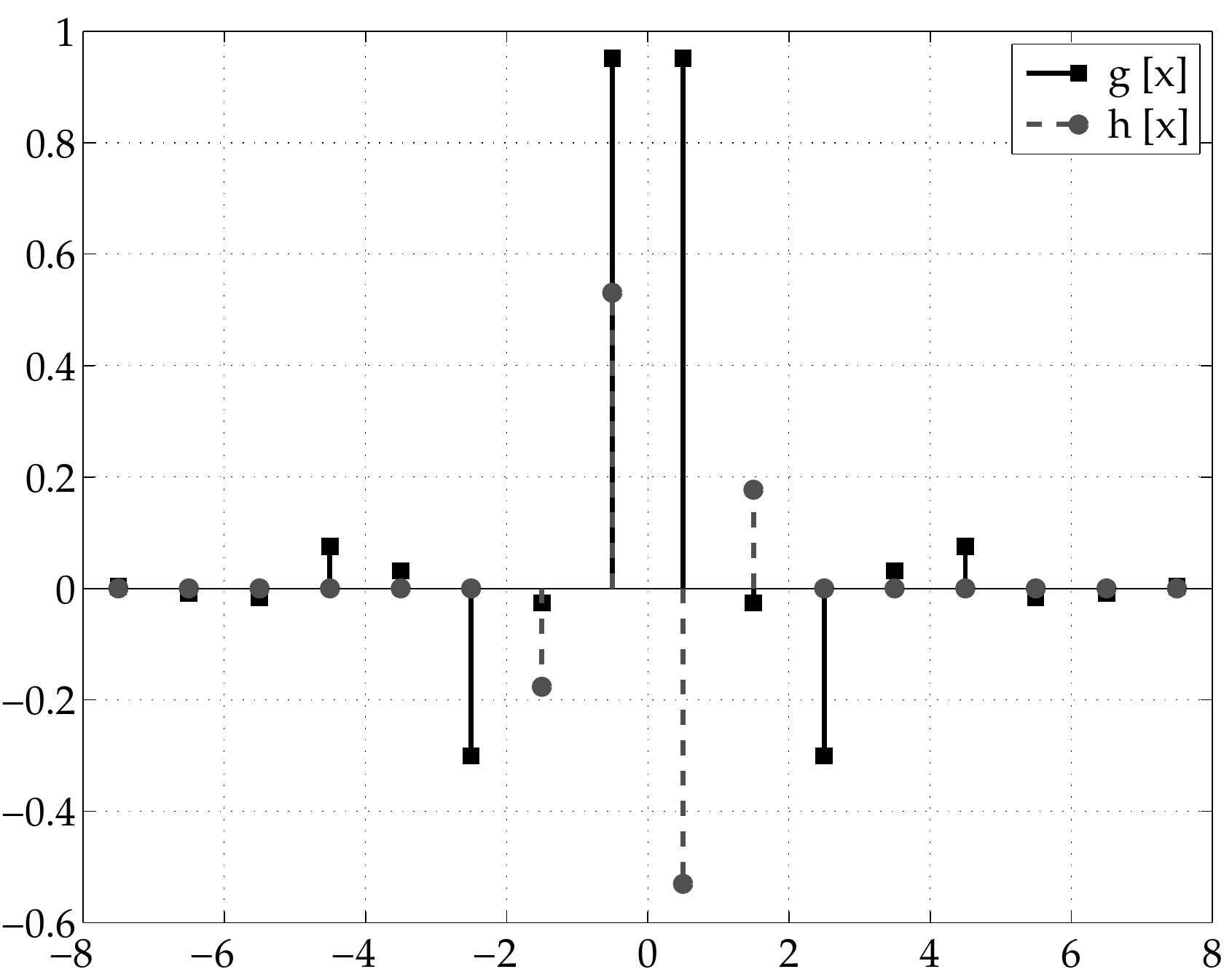}
\caption{Biorthogonal wavelet 3.7 high-pass (h) and low-pass (g) filers}
\label{fig:wletfir}
\end{minipage}
\end{figure}

For the case study proposed in this paper we have used the raw data of
connection requests over time to predict the behaviour of the users of
a widely used an internet service, i.e.\ the one provided by
Wikimedia(TM).
Raw data were taken from the page-view statistics Wikimedia(TM)
project and released by Domas Mituzas under Creative Common
License\footnote{%
  \textsf{See dumps.wikimedia.org/other/pagecounts-raw}}.  Original
data report the amount of accesses and bytes for the replies that were
sampled in time-steps of one hour for each web page accessible in the
project.
Data were collected for the whole services offered by Wikimedia
projects including Wikipedia(r), Wikidictionary(r), Wikibooks(r) and
others.
Data were gathered and composed by an automatic procedure, obtaining
the total requests made to the wikimedia servers for each hour.
Therefore, a 2-years long dataset of hourly sampled access requests
has been reconstructed. Then, this dataset was decomposed by using a
wavelet biorthogonal decomposition identified by the couple of numbers
3.7, which means that are implemented by using FIR filters with 7th
order polynomials degree for the decomposition (see also
Figure~\ref{fig:wletf} and~\ref{fig:wletfir}) and 3rd order for the
reconstruction.
Decomposed data, as shown in Figure~\ref{fig:decomp}, were then given
as inputs for the neural network as described in Section~\ref{wrnn}.
The network was trained by using a gradient descent back-propagation
algorithm with momentum led adaptive learning rate as presented
in~\cite{Haykin09}.

\begin{figure}[t]
\centering
\begin{minipage}[t]{0.49\textwidth}
\centering
\includegraphics[width=.99\textwidth]{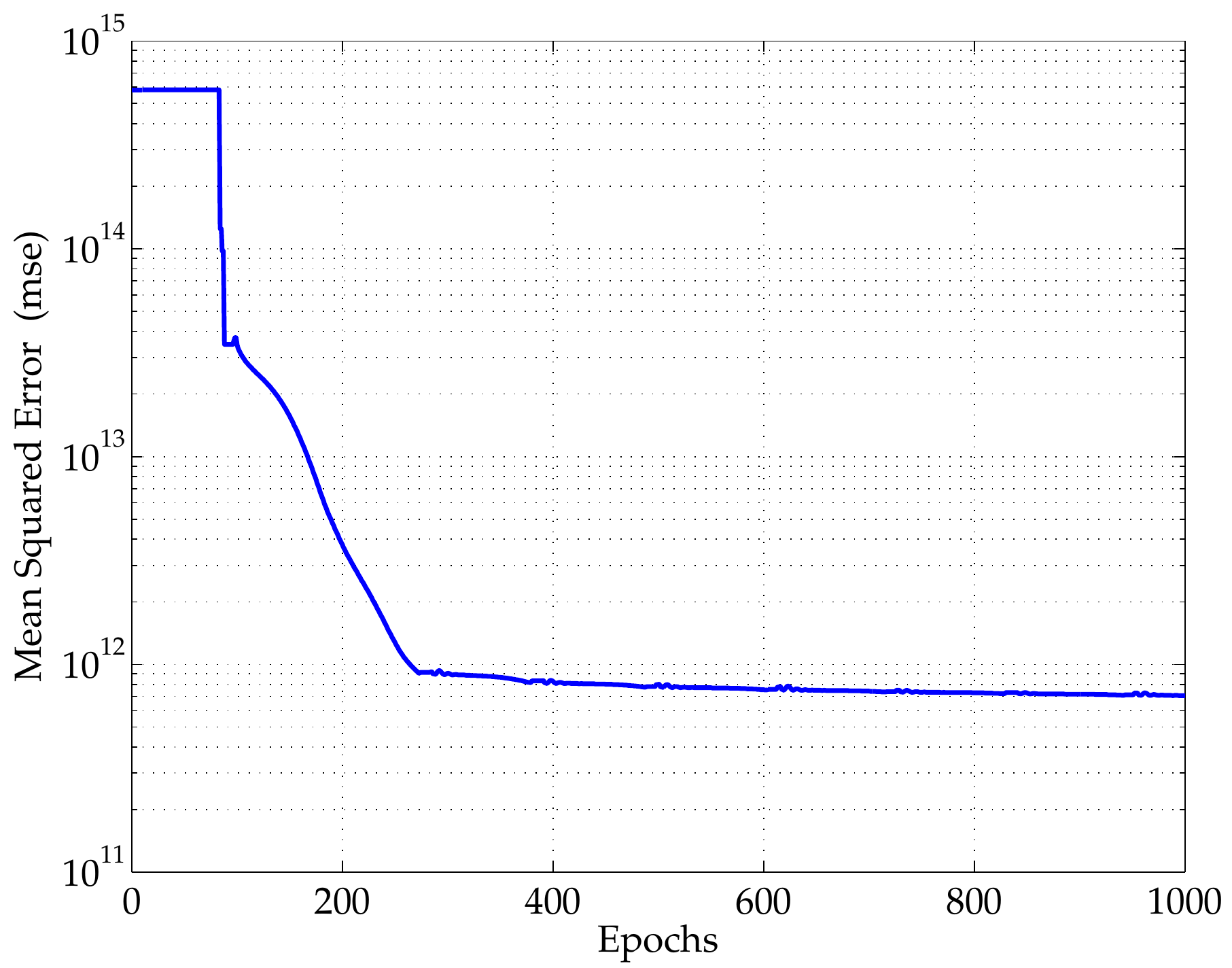}
\caption{Error while training the network} 
\label{fig:pre500}
\end{minipage}
\begin{minipage}[t]{0.49\textwidth}
\centering
\includegraphics[width=.99\textwidth]{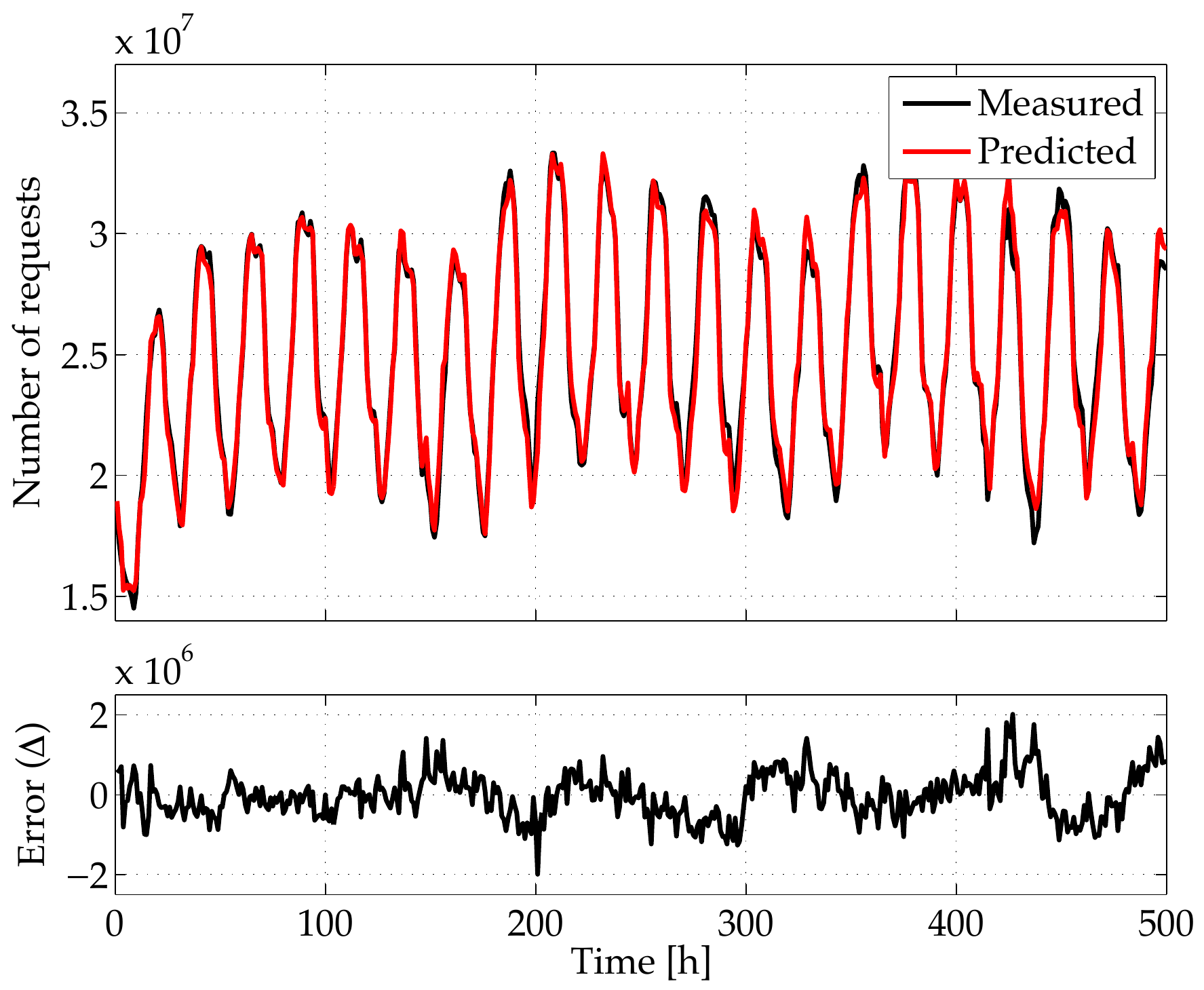}
\caption{Actual time series and predicted values}
\label{fig:nettr}
\end{minipage}
\end{figure}

\begin{figure}[t]
\centering
\centering
\includegraphics[width=.55\textwidth]{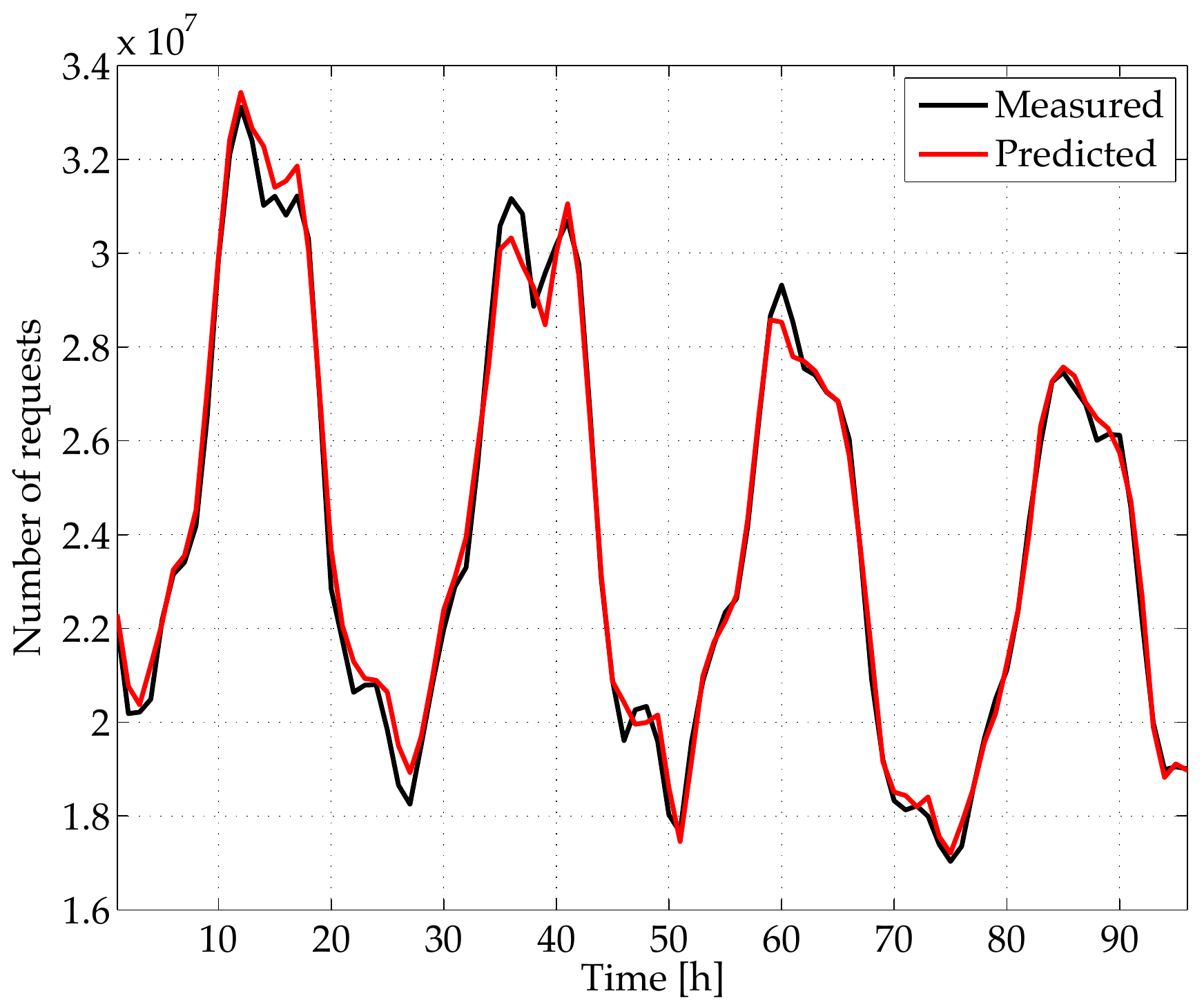}
\caption{Top of the curves for actual time series and predicted
  values, for an arbitrary chosen time-interval}
\label{fig:top}
\end{figure}

For a prediction 6 hours in advance of the time series of the amount
of requests, the root mean squared error of prediction for the access
requests over time was of $1.3156\cdot10^{04}$ requests, which means a
relative error of less than $0.6$ per thousands (less than six
requests over ten thousands).
Figure~\ref{fig:pre500} shows the shape of the mean square error while
training is being performed.
Figure~\ref{fig:nettr} shows the actual time series of the incoming
requests in black, and the predicted values for the incoming requests
in red for a time period of 500 hours. The actual values for the shown
time period had not been given as input to the neural network for
training, however have then been used to compare with the predicted
values and to compute the error (see the bottom part of
Figure~\ref{fig:nettr}).  A smaller period of time has been shown in
Figure~\ref{fig:top} to highlight the differences of actual and
predicted values. As can be seen, the neural network manages to
closely predict even relatively small variations of the trends.
The output of the neural network was then given to a resource
management service to perform allocation requests in terms of needed
bandwidth and virtual machines~\cite{ccpe13}.

\section{Conclusions}
\label{conclusions}

This paper has provided an ad-hoc architecture for a neural network
that is able to predict the amount of incoming requests performed by
users when accessing a website.
Firstly, the past time series of accesses has been analysed by means
of wavelets, which appropriately retain only the fundamental
properties of the series. Then, the neural network embeds both the
ability to perform wavelet analysis and prediction of future amount of
requests.
The performed experiments have proven that the provided ensemble is
very effective for the desired prediction, since the computed error
can be considered negligible.

Estimates can be fundamental for a resource management component, on a
server side of an internet based system, since they make it possible
to acquire just the right amount of resource (e.g.\ from a cloud).
Then, in turn it is possible to avoid an unnecessary cost and waste of
resources, whilst keeping the level of QoS as desired and unaffected
by variations of requests.

\section{Acknowledgments}

This work has been supported by project PRISMA PON04a2 A/F funded by
the Italian Ministry of University within PON 2007-2013 framework.

\bibliographystyle{abbrv}
\bibliography{nn2}

\end{document}